\documentclass[conference]{IEEEtran}
\IEEEoverridecommandlockouts
\usepackage{cite}
\usepackage{amsmath,amssymb,amsfonts}
\usepackage{algorithmic}
\usepackage{graphicx}
\usepackage{textcomp}
\usepackage{xcolor}
\usepackage[hidelinks]{hyperref}
\def\BibTeX{{\rm B\kern-.05em{\sc i\kern-.025em b}\kern-.08em
    T\kern-.1667em\lower.7ex\hbox{E}\kern-.125emX}}
\begin{document}

\title{ANCHOR: Cluster-Grounded 5W1H Extraction via Large-Small Model Collaboration}

\author{
    \IEEEauthorblockN{Hongyu Jiang, Chenang Cai, Lei Gao, Zhao Kang, Ling Tian, Jinchuan Zhang\textsuperscript{*}}
    \IEEEauthorblockA{University of Electronic Science and Technology of China, Chengdu, China\\
        Email: hyjiang@std.uestc.edu.cn, caichenang@hotmail.com, leigao@std.uestc.edu.cn,\\
        zkang@uestc.edu.cn, lingtian@uestc.edu.cn, jc.zhang@uestc.edu.cn\\
        \textsuperscript{*}Corresponding author: Jinchuan Zhang.}
}

\maketitle

\begin{abstract}
Deep semantic understanding of events requires recognizing related events and distinguishing why an event occurs from how it is carried out. 5W1H extraction provides a structured foundation for such understanding by organizing event information around six fundamental questions. However, single-event slot filling provides limited support for these distinctions. We propose \textsc{Anchor}, a cluster-grounded framework that combines a large language model (LLM) with a small language model equipped with discriminative heads for accurate 5W1H extraction. The LLM proposes candidate events and generates their initial 4W records, which guide the selection of an event cluster and the subsequent anchoring of its main event. A shared document encoder provides evidence for joint 4W refinement, and two separate sets of boundary and role-scoring heads then extract why and how under event-specific constraints, respectively. Document representations are reused across events and stages, and unsupported roles can remain empty. Following the same principles, an agent-and-skill annotation workflow constructs a benchmark of 26,110 documents from ten source corpora covering three languages and seven domains. We select nine of the resulting datasets to evaluate \textsc{Anchor} against strong baselines using four open-weight LLMs. \textsc{Anchor} achieves competitive performance. The datasets and code are available at \url{https://github.com/revolution6575/ANCHOR-5W1H}.
\end{abstract}


\section{Introduction}
\label{sec:introduction}
News articles, reports and online posts describe events in natural language. However, applications such as news aggregators, event knowledge graphs and monitoring dashboards require structured event records. Such records can also serve as structured evidence for retrieval-augmented generation (RAG) and reasoning about events and their relations. The 5W1H questions (who, when, where, what, why and how) go back to the circumstances of an action listed by Aristotle~\cite{sloan2010septem}. They break an event down into its participants, time, place, action, cause and means, and journalists answer them within the first few sentences of a story to report its main event~\cite{hamborg2018giveme5w}. They provide a compact schema for event records across domains and languages without requiring a manually constructed ontology. Therefore, extracting 5W1H information from text goes beyond language processing and constitutes a broader data engineering task. The information in a document's event records must be consistent across events and traceable to the source, including any why and how evidence that may span multiple events.

\begin{figure}[t]
	\centering
	\includegraphics[width=\columnwidth]{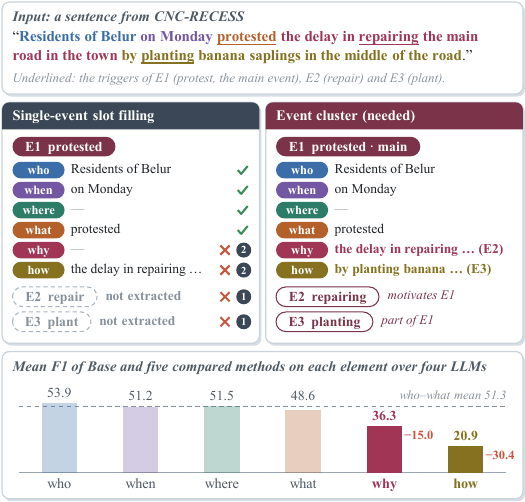}
	\caption{Challenges of 5W1H extraction on an example from CNC-RECESS. Left: a typical output of single-event slot filling, with errors numbered by challenge. Right: the event-cluster representation that the sentence calls for. Bottom: mean F1 of Base and the five LLM-based compared methods on each element, computed from the runs in Table~\ref{tab:overall-results}; for each method, the counts are pooled over the four LLMs and all nine datasets before F1 is computed.}
	\label{fig:challenges}
\end{figure}

Methods for the task have moved from hand-written rules, as in Giveme5W1H~\cite{hamborg2019giveme5w1h}, to large language models (LLMs) that are prompted~\cite{munoz2025imitating} or fine-tuned~\cite{grande2024flant5} to answer the six questions. Along the way, who, when and where have generally become easier to extract, since their answers are often short spans, usually names, dates and places. What can likewise be extracted more readily using these three roles as context, whereas why and how often remain more challenging. Why and how were answered least precisely in the human evaluation of Giveme5W1H, while ChatGPT (GPT-3.5-turbo)~\cite{ouyang2022instructgpt} struggled most with what, why and how in the study by Cao et al.~\cite{cao2024fivew}. Five-shot LLMs evaluated on EVENT5Ws~\cite{event5ws} achieve an exact-match F1 of up to 63.3 for when but no more than 11.3 for why, while how is absent from the benchmark's annotations and evaluation. A directly prompted LLM (Base) and the five LLM-based methods compared in this paper show the same gap on our benchmark, with a mean F1 of 36.3 on why and 20.9 on how, against 48.6 to 53.9 on the other four elements (Fig.~\ref{fig:challenges}).

Extracting why and how remains challenging because of limitations in event representation and role differentiation, as well as a shortage of complete 5W1H frameworks and annotated datasets. Consider a sentence from the CNC-RECESS~\cite{tan2023recess} corpus: \textit{``Residents of Belur on Monday protested the delay in repairing the main road in the town by planting banana saplings in the middle of the road.''} The phrase \textit{the delay in repairing the main road in the town} explains \textit{why} the residents protested and describes the state of a separate repair event. The phrase \textit{by planting banana saplings in the middle of the road} describes a third event. Planting the saplings was part of the protest and explains \textit{how} the residents carried it out. Answers to why and how are usually clauses or phrases and often refer to other events related to the target event. An answer to why often refers to another event or state that causes or motivates the target event, so its extraction requires identifying the explanatory relation between them. An answer to how describes the means or procedure through which the target event is carried out and must be interpreted in relation to that event's action and arguments. A span can therefore answer why for one event without filling any role for another. Unlike names or dates, answers to why and how are not readily obtained through single-event slot filling. Instead, why and how must be determined separately for each target event, which may require linking events within a document and distinguishing these two roles from the remaining four roles. Yet some existing 5W1H methods~\cite{hamborg2019giveme5w1h,munoz2025imitating} still focus on answering the six questions for a single main event, making these requirements difficult to satisfy. These limitations give rise to three challenges in 5W1H extraction, and Fig.~\ref{fig:challenges} illustrates the first two on this example.

\textbf{Challenge 1:} Keeping one event in view loses the links between events. 5W1H systems and benchmarks describe a document by its main event~\cite{hamborg2019giveme5w1h,event5ws}. In event argument extraction (EAE), RAMS~\cite{ebner2020rams} annotates a target event within a five-sentence window around its trigger, and DocEE~\cite{docee2022} keeps one main event per document. Many EAE methods also process one given trigger at a time~\cite{ma2022paie,code4struct2023,hdloa2024,cat2025}. Extracting only the protest in the example drops the repair and planting events, together with the links showing that the delay motivated the protest and that the planting carried it out.

\textbf{Challenge 2:} Treating why and how as the same type of slot overlooks their different functions. Why gives a cause or purpose of an event, whereas how gives the means or procedure by which it is carried out. Schema-based EAE assigns spans to predefined argument roles~\cite{doddington2004ace}, and event causality identification (ECI) only decides whether two events are causally related~\cite{ecisurvey2025}. Neither locates the span that answers why or how for a target event.

\textbf{Challenge 3:} Frameworks and annotated datasets for complete 5W1H extraction remain limited. Some systems and benchmarks~\cite{hamborg2018giveme5w,event5ws} cover only the five W questions and omit how. Fine-tuned extractors require annotated training data~\cite{cao2024fivew}, and general-purpose EAE methods such as PAIE~\cite{ma2022paie} and Code4Struct~\cite{code4struct2023} were not designed for 5W1H extraction. The available 5W1H datasets are mostly monolingual English or Spanish news~\cite{hamborg2019giveme5w1h,cao2024fivew,flares2024}, which leaves evaluation across languages and domains unsupported.

We address these limitations with \textsc{Anchor}, a cluster-grounded framework that combines a large language model with a small language model (SLM) equipped with discriminative heads for complete 5W1H extraction. The selected cluster provides the context for assigning source spans to the 5W1H roles of each retained event. The LLM first proposes multiple candidate events and generates their 4W drafts. For each candidate, a single response first extracts who, when and where, then generates what from the document and these preceding participant, temporal and spatial cues. These drafts inform the selection of an event cluster, after which the main event is identified within the cluster and the identities of its events remain fixed. A shared document encoder then proposes role-specific evidence to guide the LLM in jointly refining who, when, where and what. Conditioned on the refined 4W, separate boundary and role-scoring heads extract and score why and how. Their scores combine role evidence with boundary support, and constrained selection allows unsupported roles to remain empty to avoid forcing answers when the document provides no supporting evidence. The document representations are reused across events and stages, and why/how selection requires no further LLM calls. We also apply the same principles of event-cluster organization, event-specific role assignment and source-grounded evidence in an agent-and-skill workflow for reconstructing dataset annotations.

Our benchmark comprises 26,110 documents from ten source corpora, combining preserved human 5W1H annotations with annotations reconstructed through our workflow. The documents span Chinese, English and Spanish and cover the news, health, medical, emergency, finance, military and general domains. Experiments show that \textsc{Anchor} achieves competitive 5W1H extraction performance against the evaluated baselines.

Our contributions are as follows.
\begin{itemize}
	\item We propose an event-cluster extraction strategy that groups the selected events of a document into a cluster and anchors the main event within it. Each event in the cluster has its own source-grounded 5W1H record, providing an explicit anchor for assigning information to the six roles.
	\item We develop separate extraction and scoring mechanisms for why and how. Independent boundary and role-scoring heads use the refined 4W as event context, combine role-specific semantic evidence with boundary support, and select supported spans under role-specific thresholds and within-event constraints.
	\item We integrate the LLM and the SLM into a complete 5W1H extraction pipeline through shared document encoding, evidence-guided 4W refinement and discriminative selection of why and how answers. The resulting pipeline achieves competitive performance against the evaluated baselines. Following the same principles, we use agents equipped with reusable extraction and review skills to reconstruct 5W1H annotations for a benchmark of 26,110 documents from ten source corpora across three languages and seven domains while preserving existing human 5W1H labels.
\end{itemize}

\section{Related Work}
\label{sec:related-work}

\subsection{5W1H Extraction}
\label{sec:5w1h-extraction}

Hamborg et al.~\cite{hamborg2018giveme5w} built one of the earliest open-source 5W extraction systems for news articles. It is syntax-based and answers five of the six 5W1H questions (all but how). Hamborg et al. later extended it to answer how as well, yielding Giveme5W1H~\cite{hamborg2019giveme5w1h}. Its syntactic and domain-specific rules propose candidate phrases for each question, and a weighted score over factors such as position and frequency in the article picks the best one.

Cao et al.~\cite{cao2024fivew} applied LLMs to the task: they annotated 5W1H answers for articles from four English news corpora (CNN/DailyMail, XSum, NYT and RA-MDS) and found that open-source LLMs fine-tuned on these annotations outperform zero-shot and few-shot ChatGPT. Sharma et al.~\cite{event5ws} later released a larger benchmark named EVENT5Ws that marks the five Ws without how for one event in each of its 10,000 news documents from the title and the first five sentences. Sep\'{u}lveda-Torres et al.~\cite{flares2024} set up a subtask on identifying 5W1H elements in Spanish news within the FLARES shared task at IberLEF. Cassola-Bacallao et al.~\cite{llmjudge5w1h2026} then used LLMs to judge 5W1H extractions on FLARES and another Spanish corpus and found that the judgments largely agree with those of journalism experts.

\subsection{Task-Specific Models for Event Argument Extraction}
\label{sec:event-argument-extraction}

Event argument extraction grew out of the Automatic Content Extraction (ACE) program~\cite{doddington2004ace}, which defined a set of event types and the argument roles of each type. Lin et al.~\cite{oneie2020} extracted entities, triggers, relations and arguments jointly as a graph. Du and Cardie~\cite{eeqa2020} cast argument extraction as question answering with one question per role in EEQA. These two methods work within a sentence, whereas the arguments of an event often span several sentences, and this led to document-level work on cross-sentence arguments. This shift began with datasets that annotate arguments beyond a single sentence. Ebner et al.~\cite{ebner2020rams} built RAMS with a five-sentence window around each trigger. Li et al.~\cite{li2021wikievents} annotated full articles in WikiEvents. Tong et al.~\cite{docee2022} built DocEE on full articles as well. Zhu et al.~\cite{cmnee2024} annotated Chinese military news at the document level in CMNEE. On the model side, Ma et al.~\cite{ma2022paie} prompted all the roles of an event together instead of asking one question per role as EEQA does. To use context beyond the trigger sentence, Xu et al.~\cite{tsar2022} linked information across sentences through AMR graphs, and Zhang et al.~\cite{gam2024} through a semantic mention graph. These methods still extract the arguments of one event at a time. To address this, Liu et al.~\cite{liu2024deeia} appended the prompts of all events to the document and extracted their arguments in one pass.

Later work widened what counts as an argument. Wang et al.~\cite{wang2024mavenarg} built MAVEN-Arg, in which an argument can be any span of text, not just an entity. In DiscourseEE, Sharif et al.~\cite{discourseee2024} further added implicit arguments that can only be inferred from context and scattered arguments that are spread across the text. Beyond the arguments of a single event, the relations between events have also been studied. Wang et al.~\cite{mavenere2022} labeled causal, temporal, subevent and coreference relations between events in MAVEN-ERE.

\subsection{Large Language Models for Event Argument Extraction}
\label{sec:llm-event-argument-extraction}

Recently, LLMs have also been applied to EAE, as they can extract arguments without being trained for the task. Huang et al.~\cite{textee2024} re-evaluated 14 methods on 16 datasets with the same preprocessing and data splits and found that LLMs prompted directly with a few demonstrations fall far behind much smaller models fine-tuned on annotated data. Direct prompting thus leaves much room for improvement. Many methods work in this direction without training the LLM, either recasting extraction as another task or adding intermediate steps before the answer. Wei et al.~\cite{chatie2023} cast zero-shot extraction as multi-turn question answering in ChatIE. In Code4Struct, Wang et al.~\cite{code4struct2023} had the LLM write each event as an instance of a Python class. In HD-LoA, Zhou et al.~\cite{hdloa2024} prompted the LLM to reason by analogy from demonstrations built on task heuristics. Zhang et al.~\cite{ultra2024} let the LLM read a document chunk by chunk to collect candidate arguments and then drop the irrelevant ones through self-refinement. In CAT, Wei et al.~\cite{cat2025} asked the LLM to reason about the event before choosing each argument from a set of candidates. These methods narrow the gap, and some of them reach or even surpass fine-tuned models on certain benchmarks~\cite{code4struct2023,ultra2024}.

With stronger LLMs, more recent systems go further and divide the work among multiple agents. Luo et al.~\cite{oneke2025} built OneKE with separate agents for schema construction, extraction and reflection. Guo et al.~\cite{aec2026} built AEC, in which a verification agent checks each output against the event schema written as an executable class.

\section{Problem Formulation}
\label{sec:problem-formulation}
\begin{figure*}[tbp]
\centering
\includegraphics[width=\textwidth]{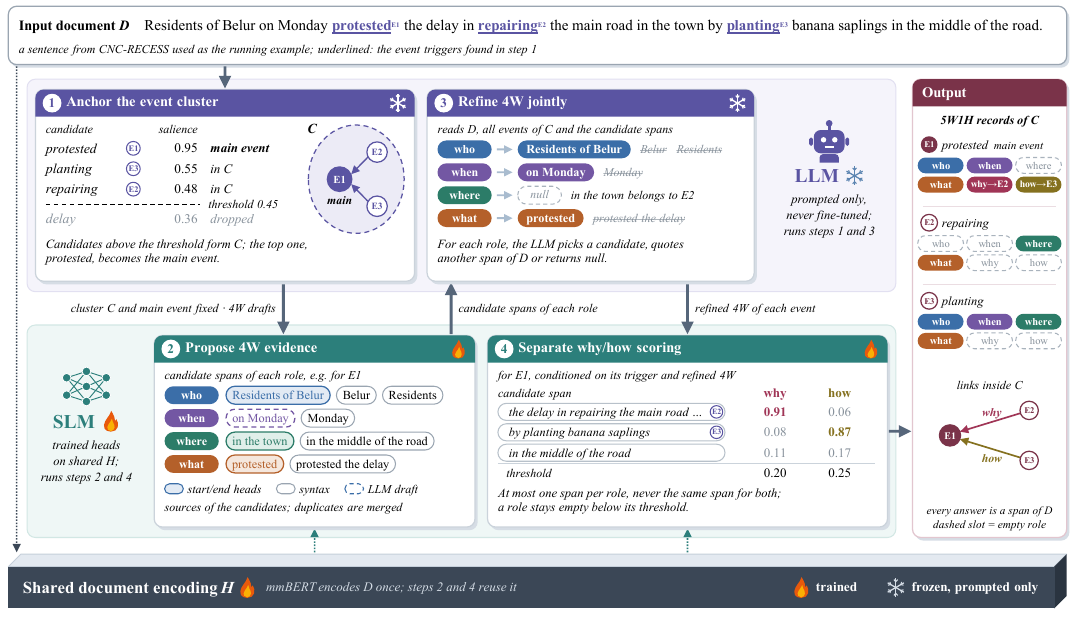}
\caption{Framework of \textsc{Anchor} for 5W1H extraction.}
\label{fig:anchor-framework}
\end{figure*}

Let $D$ be a document in Chinese, English or Spanish, and let $\mathcal{S}(D)$ be the set of its contiguous text spans. An event $e$ in $D$ is located by a trigger span $t_e\in\mathcal{S}(D)$ and described by six roles $\mathcal{R}=\mathcal{R}_4\cup\mathcal{R}_2$, where $\mathcal{R}_4=\{\textit{who},\textit{when},\textit{where},\textit{what}\}$ and $\mathcal{R}_2=\{\textit{why},\textit{how}\}$. Given $D$ alone, the task is to output
\begin{equation}
\mathcal{Y}=\bigl(\mathcal{C},\,e^{*},\,\{A_e\}_{e\in\mathcal{C}}\bigr),\qquad A_e(r)\subseteq\mathcal{S}(D)\ \ (r\in\mathcal{R}),
\label{eq:task-output}
\end{equation}
where the event cluster $\mathcal{C}$ is a set of related events reported in $D$, $e^{*}\in\mathcal{C}$ is its main event and $A_e$ is the 5W1H record of event $e$. An empty $A_e(r)$ means that $D$ does not answer role $r$ for $e$. Since every answer is a span of $D$, each record can be traced back to the source. Existing 5W1H systems output the record of the main event only, which is the special case $\mathcal{C}=\{e^{*}\}$. Multi-event argument extraction~\cite{liu2024deeia} receives the triggers of all events and treats them as a set without a main event.

\section{Methodology}
\label{sec:methodology}
\textsc{Anchor} builds $\mathcal{Y}$ in four steps that alternate between an LLM and an SLM with discriminative heads (Fig.~\ref{fig:anchor-framework}). The LLM handles the steps that need the whole document in view. It proposes candidate events with 4W drafts, from which the event cluster is selected and its main event anchored (Section~\ref{sec:anchoring}), and it later refines the 4W of each selected event (Section~\ref{sec:llm-refinement}). The SLM scores spans on a shared encoding of the document. It proposes the evidence used in the refinement (Section~\ref{sec:slm-evidence}) and extracts why and how with separate heads (Section~\ref{sec:why-how}). The LLM is only prompted and never fine-tuned, and the SLM is trained as described in Section~\ref{sec:training}.

\subsection{Selecting the Event Cluster and Anchoring the Main Event}
\label{sec:anchoring}
\textsc{Anchor} drafts the 4W of each candidate before selecting events, so that the evidence found for a candidate counts toward its selection. The LLM first proposes up to five candidate events, each given by a trigger, the sentence containing it and its relations to the other candidates. Triggers that cannot be located in $D$ or that overlap an earlier trigger are dropped. For each remaining candidate, one LLM request drafts its who, when, where and what together, and the answers are mapped to spans of $D$ to form the draft record $A_e^{0}$. The candidates are then ranked by
\begin{equation}
\mathrm{Sal}(e)=w^{\top}\bigl(\mathrm{Pos}(e),\,\mathrm{Trig}(e),\,\mathrm{Cov}(e),\,\mathrm{Cent}(e)\bigr),
\label{eq:salience}
\end{equation}
where $w$ holds four fixed weights, $\mathrm{Pos}$ favors triggers in the opening sentences, $\mathrm{Trig}$ indicates whether the trigger is the main predicate of its sentence, $\mathrm{Cov}$ is the fraction of the four roles with a grounded draft and $\mathrm{Cent}$ grows with the number of relations pointing to $e$ from other candidates. The cluster $\mathcal{C}$ keeps up to three candidates whose salience reaches a threshold, and the highest-ranked one becomes the main event $e^{*}$. The candidate limit of five matches the maximum number of events per annotated cluster in the datasets listed in Table~\ref{tab:statistics-of-the-datasets}, while retaining at most three salient events controls the output size. Links from the repair and planting events contribute to the protest's $\mathrm{Cent}$ score in Fig.~\ref{fig:anchor-framework}. The protest ranks first, and both related events remain in $\mathcal{C}$. The members of $\mathcal{C}$ and the choice of $e^{*}$ are fixed after this step, so the later steps fill roles without redefining the events.

\subsection{Evidence Preparation with the SLM}
\label{sec:slm-evidence}
A single LLM pass can miss a role or return text that does not match the source. Preparing refinement evidence with the LLM can also require repeated calls and increase inference time. \textsc{Anchor} therefore uses evidence from the SLM to refine the drafts.
The multilingual encoder mmBERT~\cite{marone2026mmbert} reads $D$ in 512-token windows that overlap by 128 tokens, and the states of tokens covered by more than one window are averaged into $H\in\mathbb{R}^{L\times d}$ for the $L$ tokens of $D$. This encoding is computed once per document and reused for every event and for the extraction of why and how. For an event $e$, a query vector $q_e$ combines the pooled states of its trigger and its sentence with a flag marking the main event. Each role $r\in\mathcal{R}_4$ has its own start and end heads, $\pi_r^{\mathrm{s}}(i\mid e)=\sigma(g_r^{\mathrm{s}}(H_i,q_e))$ and $\pi_r^{\mathrm{e}}(i\mid e)=\sigma(g_r^{\mathrm{e}}(H_i,q_e))$, where $\sigma$ is the sigmoid function and $g_r^{\mathrm{s}}$ and $g_r^{\mathrm{e}}$ are feed-forward scorers. A span $c$ from token $i_c$ to token $j_c$ receives the boundary support
\begin{equation}
B_r(e,c)=\sqrt{\pi_r^{\mathrm{s}}(i_c\mid e)\,\pi_r^{\mathrm{e}}(j_c\mid e)}.
\label{eq:boundary}
\end{equation}
The candidate list of each role holds the spans with the highest support, a few spans taken from the syntax of $D$ and the draft answer, with duplicates merged.

\subsection{LLM Refinement Guided by SLM Evidence}
\label{sec:llm-refinement}
The LLM then refines the four roles of each event jointly in one request. The prompt shows the numbered sentences of $D$, the triggers of all events in $\mathcal{C}$, the target event and the candidate lists. Each candidate appears with an identifier, its text and its sentence, while the boundary support scores $B_r(e,c)$ computed by the SLM are omitted. For each role, the LLM picks a candidate, quotes another span of $D$ together with its sentence, or returns null, after which the answer is checked against $D$. The SLM proposes candidate spans, and the LLM uses the selected cluster and the full document to assign 4W answers to the target event. The result is the refined 4W record $A_e^{\star}$.

\subsection{Separate Extraction of Why and How}
\label{sec:why-how}
Why and how ask different questions about the same event, so \textsc{Anchor} uses the SLM to extract them with separate boundary heads and scoring functions. Both are conditioned on the event through the vector
\begin{equation}
u_e=\mathrm{MLP}_E\bigl[\mathrm{Pool}(H,t_e);\,[\mathrm{Pool}(H,A_e^{\star}(r))]_{r\in\mathcal{R}_4};\,m_e\bigr],
\label{eq:event-vector}
\end{equation}
where $\mathrm{Pool}$ averages the rows of $H$ covered by a span, $[\,;\,]$ denotes concatenation, an empty role contributes a zero vector and $m_e\in\{0,1\}^4$ marks the nonempty roles. The vector tells the heads which action or state a candidate has to explain, and it is computed from $H$ without encoding the document again.

The candidate pool $\mathcal{P}_e$ merges the top spans of the why and how boundary heads, which are conditioned on $u_e$, with clauses, predicate-centered phrases and cue-bearing phrases taken from the syntax of $D$. Spans that express other events of $\mathcal{C}$ stay in the pool so that a related event can answer why or how for $e$, with only the trigger of $e$ excluded. A candidate $c$ is represented by $v_c$, built from the states at its two ends, its pooled states and its length. For $r\in\mathcal{R}_2$, the role logit is
\begin{equation}
\begin{split}
z_r(e,c)={}&(W_r u_e)^{\top}(V_r v_c)/\sqrt{k}\\
&+\mathrm{MLP}_r\bigl[u_e;v_c;u_e\odot v_c;|u_e-v_c|;f_r(e,c)\bigr],
\end{split}
\label{eq:role-logit}
\end{equation}
where $W_r$ and $V_r$ project to rank $k$ and $f_r$ holds role-specific features. The why features cover causal cues, sentence distance and temporal order. The how features cover subordination or shared arguments, manner cues, sentence distance and causal cues. Since why and how have separate parameters and separate sigmoids rather than one softmax over the roles, each of them can be present or absent independently. The semantic score is combined with the boundary support of Eq.~\eqref{eq:boundary}, here computed with the why or how heads:
\begin{equation}
S_r(e,c)=\sigma\bigl(z_r(e,c)/T_r\bigr)\bigl[(1-\lambda_r)+\lambda_r B_r(e,c)\bigr],
\label{eq:gated-score}
\end{equation}
where $T_r>0$ is a temperature and $\lambda_r\in[0,1]$ weights the boundary support. Weak support lowers the score of a candidate, but even full support cannot raise it above its semantic score.

\textsc{Anchor} then assigns at most one span to each role by solving
\begin{equation}
\begin{aligned}
\max_{x}\ &\textstyle\sum_{c\in\mathcal{P}_e}\sum_{r\in\mathcal{R}_2}x_{c,r}\bigl(S_r(e,c)-\tau_r\bigr)\\
\text{s.t.}\ &\textstyle\sum_{c}x_{c,r}\le1,\quad x_{c,\textit{why}}+x_{c,\textit{how}}\le1,
\end{aligned}
\label{eq:selection}
\end{equation}
where $x_{c,r}\in\{0,1\}$ indicates that $c$ fills role $r$ and $\tau_r$ is a role-specific threshold. A candidate that scores no higher than $\tau_r$ cannot increase the objective, so a role stays empty when no candidate exceeds its threshold. The second constraint keeps one span from being both the why and the how of the same event. In conventional joint extraction, why and how can be conflated, leading to the same span being assigned to both roles. With only two roles, the optimum is found by enumerating the two best spans of each role and the empty choice.

\begin{figure*}[tbp]
\centering
\includegraphics[width=\textwidth]{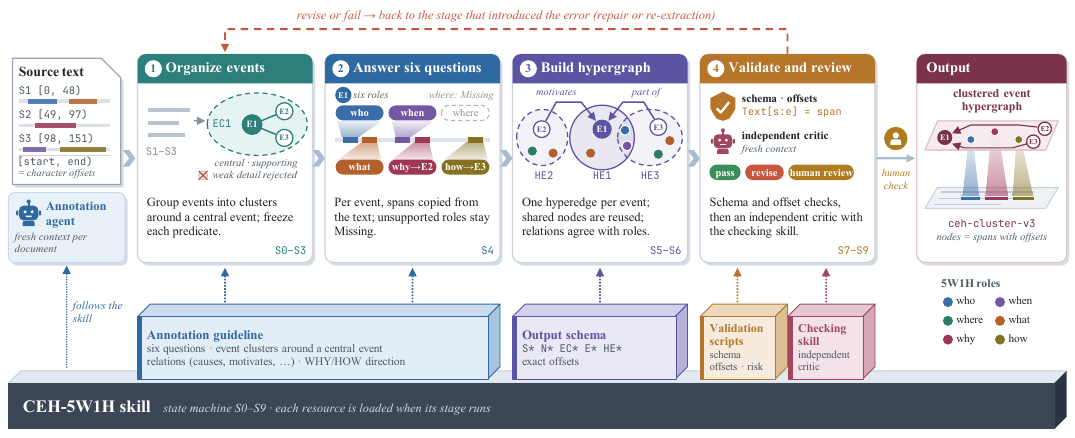}
\caption{Framework of the CEH-5W1H skill for 5W1H annotation. CEH stands for Clustered Event Hypergraph.}
\label{fig:ceh-skill}
\end{figure*}

\subsection{Training and Inference}
\label{sec:training}
The SLM is trained on 722 annotated documents obtained by randomly sampling 10\% of each of the nine comparison datasets, with per-dataset counts rounded up. Sampling is stratified by the presence of why and how, and the selected subset also supports hyperparameter selection. Predicted and gold events are matched one to one by trigger and what span overlap. For why and how, candidates are positive at character-level IoU $\ge0.5$ with a gold answer, negative when they overlap no gold answer, and ignored otherwise. The objective combines class-balanced binary cross-entropy for role scoring and boundary prediction with a margin ranking loss, weighted 1, 0.5 and 0.2, respectively. The 4W heads are trained afterwards with the encoder frozen. The salience weights in Eq.~\eqref{eq:salience} are $(0.30,0.30,0.25,0.15)$, with a threshold of 0.45. The parameters $(T_r,\lambda_r,\tau_r)$ are selected using out-of-fold predictions and shared across all three languages: $(0.5,0.25,0.20)$ for why and $(2,0.5,0.25)$ for how. For comparison with the LLM baselines on a common document set, we evaluate all 7,174 documents from these datasets, including the 722 training and calibration documents.

At inference time, a document with $P\le5$ grounded candidates costs $1+P+|\mathcal{C}|$ LLM requests: one for the proposal, one draft per candidate and one refinement per selected event. The proposal and draft prompts list at most the first 40 sentences of $D$, and the refinement prompt lists all of them. Why and how add no request, and each document is encoded once.

\subsection{Consistency with the Benchmark Annotation}
\label{sec:benchmark-consistency}
For the mapped and agent-extracted groups in Section~\ref{sec:datasets}, the annotation agents follow the same principle as \textsc{Anchor} by assigning source spans to the roles of individual events within a cluster. The extraction and review agents follow the CEH-5W1H skill (Fig.~\ref{fig:ceh-skill}), where CEH stands for Clustered Event Hypergraph. It packages the annotation guideline, the output schema and the validation scripts. The skill organizes annotation as a state machine (S0--S9) that advances from event organization and role extraction to hypergraph construction, validation and review. The guideline defines the six questions, how the events of a document form clusters around a central event and relations such as causes and motivates. The extraction agent groups the events of a document into clusters and revises mapped candidates when available. It then answers the six questions for each event with spans copied from the text, leaving unsupported roles empty. The annotations form a clustered event hypergraph, where each event is represented by a hyperedge connecting its available 5W1H span nodes. Nodes are reused across events when the same referent fills the same role, and directed relations connect related events. An independent review agent checks event organization, role assignments and consistency with the source text. Errors detected during validation or review return the workflow to the stage that introduced them for correction and revalidation. This division is intended to balance annotation efficiency with careful semantic verification.

\section{Experiments}
\label{sec:experiments}
\textsc{Anchor} is compared with six representative 5W1H and event extraction methods on nine datasets, and every LLM-based method runs on a unified set of four open-weight LLMs. We evaluate only Base on DEIE because each method takes over seven days per LLM. Our contribution on this corpus is the reconstruction of its 5W1H annotations.

\subsection{Experimental Settings}
\label{sec:experimental-settings}

\subsubsection{Datasets}
\label{sec:datasets}
The experiments use ten 5W1H datasets, each built from a public corpus in Chinese, English or Spanish. Table~\ref{tab:statistics-of-the-datasets} gives their statistics.

\begin{table*}[htbp]
	\renewcommand{\arraystretch}{1.2}
	\caption{Statistics of the Datasets}
	\label{tab:statistics-of-the-datasets}
	\begin{center}
		\begin{tabular*}{\textwidth}{@{\extracolsep{\fill}}lccrr@{\hskip 14pt}rrrrrr@{}}
			\hline\noalign{\vskip 1pt}
			\textbf{Dataset} & \textbf{Lang.} & \textbf{Domain} & \textbf{Docs} & \textbf{Events} & \multicolumn{6}{c}{\textbf{Element coverage of events (\%)}} \\
			\cline{6-11}\noalign{\vskip 1.4pt}
			& & & & & \textbf{Who} & \textbf{When} & \textbf{Where} & \textbf{What} & \textbf{Why} & \textbf{How} \\
			\hline\noalign{\vskip 1pt}
			\multicolumn{11}{l}{\textit{Original}} \\
			Giveme5W1H & en & News & 96 & 96 & 100.00 & 83.33 & 75.00 & 98.96 & 84.38 & 87.50 \\
			FLARES & es & News & 1,582 & 1,582 & 66.31 & 37.86 & 35.46 & 86.16 & 13.40 & 28.76 \\
			\hline\noalign{\vskip 1pt}
			\multicolumn{11}{l}{\textit{Mapped}} \\
			CNC-RECESS & en & News & 1,807 & 2,904 & 82.44 & 35.37 & 34.47 & 100.00 & 59.19 & 7.71 \\
			DiscourseEE & en & Health & 312 & 598 & 93.14 & 28.09 & 3.51 & 100.00 & 42.47 & 51.17 \\
			PHEE & en & Medical & 28 & 34 & 79.41 & 20.59 & 0.00 & 100.00 & 64.71 & 47.06 \\
			FrameNet & en & General & 13 & 18 & 100.00 & 0.00 & 27.78 & 100.00 & 83.33 & 72.22 \\
			CEC & zh & Emergency & 332 & 756 & 66.80 & 68.92 & 72.62 & 100.00 & 36.77 & 21.43 \\
			OEE-CFC & zh & Finance & 7 & 29 & 89.66 & 24.14 & 17.24 & 100.00 & 55.17 & 51.72 \\
			DEIE & zh & News & 18,936 & 69,696 & 85.55 & 72.90 & 52.18 & 100.00 & 56.88 & 27.63 \\
			\hline\noalign{\vskip 1pt}
			\multicolumn{11}{l}{\textit{Agent-Extracted}} \\
			CMNEE & zh & Military & 2,997 & 6,573 & 93.60 & 52.59 & 32.31 & 100.00 & 30.14 & 23.02 \\
			\hline
		\end{tabular*}
	\end{center}
\end{table*}

\begin{itemize}
	\item Giveme5W1H~\cite{hamborg2019giveme5w1h}: English news articles annotated by hand for all six questions.
	\item FLARES~\cite{flares2024}: Spanish news with hand-annotated 5W1H elements, from an IberLEF shared task.
	\item CNC-RECESS~\cite{tan2023recess}: English news sentences that each express a causal relation, with the cause, effect and signal spans marked.
	\item DiscourseEE~\cite{discourseee2024}: Posts and comments from English online health discussions, whose event arguments include implicit ones inferred from context and scattered ones spread across the text.
	\item PHEE~\cite{sun2022phee}: English sentences from medical case reports, annotated with adverse and therapeutic drug events and their arguments.
	\item FrameNet~\cite{baker1998framenet}: English sentences annotated with semantic frames and their frame elements.
	\item CEC~\cite{cec_corpus}: Chinese news on emergencies, annotated with event triggers, participants, objects, time and location, which map closely onto who, when, where and what.
	\item OEE-CFC~\cite{wan2024oeecfc}: Chinese financial commentaries annotated for open event extraction.
	\item DEIE~\cite{deie2024}: Chinese news with document-level annotation of events and their arguments.
	\item CMNEE~\cite{cmnee2024}: Chinese military news with document-level event annotation, in which one article often reports several events.
\end{itemize}

Altogether, 82,286 events are annotated in 26,110 documents. CMNEE, PHEE, FrameNet and OEE-CFC are used in part, with priority given to documents that explicitly describe the cause of an event or the means or procedure through which it is carried out. Duplicate documents are removed from all datasets.

The ten datasets fall into three groups according to how their 5W1H annotations are built. This design preserves human labels that already follow the 5W1H schema and reuses other structured annotations where they can be mapped to it. Direct extraction from source text provides a complementary setting in which the agents construct complete event representations without relying on imported labels. In the \textit{original} group, the human 5W1H annotations of Giveme5W1H and FLARES are kept unchanged. The other two groups are annotated by LLM agents. In both of them, the annotations are then checked by an independent agent and finally by hand. The annotation and review procedures are detailed in the following subsection.

\subsubsection{Annotation Agents}
\label{sec:annotation-agents}
GPT-5.6-Sol performs extraction, and GPT-6-Astra performs independent review in the Codex harness. During extraction, the GPT-5.6-Sol agent processes each document in a fresh context. Each output is checked against the schema and the character offsets, and failed documents are sent back for repair or re-extraction. The GPT-6-Astra review agent then checks every annotated document with a separate checking skill. Finally, the annotations are checked by hand.

In the \textit{mapped} group, the cause--effect, event and frame annotations of CNC-RECESS, DiscourseEE, PHEE, FrameNet, CEC, OEE-CFC and DEIE are first mapped by rules to candidate answers. In CNC-RECESS, for example, a cause span becomes a candidate for why and an effect span a candidate for what. An annotation agent then checks every candidate against the full text and keeps it, shortens it, drops it or moves it to the question it actually answers. For the questions that no candidate covers, such as who and when in CNC-RECESS, the agent starts from the event and the answers kept so far and extracts the remaining answers of that event from the text. In the \textit{agent-extracted} group, the events and 5W1H answers of CMNEE are extracted entirely by the agents.

\subsubsection{Compared Methods}
\label{sec:compared-methods}

We do not fine-tune the backbone LLMs for 5W1H extraction in any of the evaluated methods. They fall into two groups by what they receive besides the document.
\begin{itemize}
	\item Document only: Giveme5W1H~\cite{hamborg2019giveme5w1h}, ChatIE~\cite{chatie2023} and OneKE~\cite{oneke2025}. Like \textsc{Anchor}, they have to find the events themselves. Giveme5W1H is the only compared method that runs without an LLM.
	\item Gold triggers: Code4Struct~\cite{code4struct2023}, HD-LoA~\cite{hdloa2024} and CAT~\cite{cat2025}. As event argument extraction methods, they start from a given event trigger and cannot run without one. They receive the trigger of each gold event and answer its six 5W1H questions: who, when, where, what, why and how. HD-LoA reasons by analogy from demonstrations and is given one demonstration in addition to the triggers, as required by the original method.
\end{itemize}

\begin{figure}[htbp]
\centering
\includegraphics[width=\columnwidth]{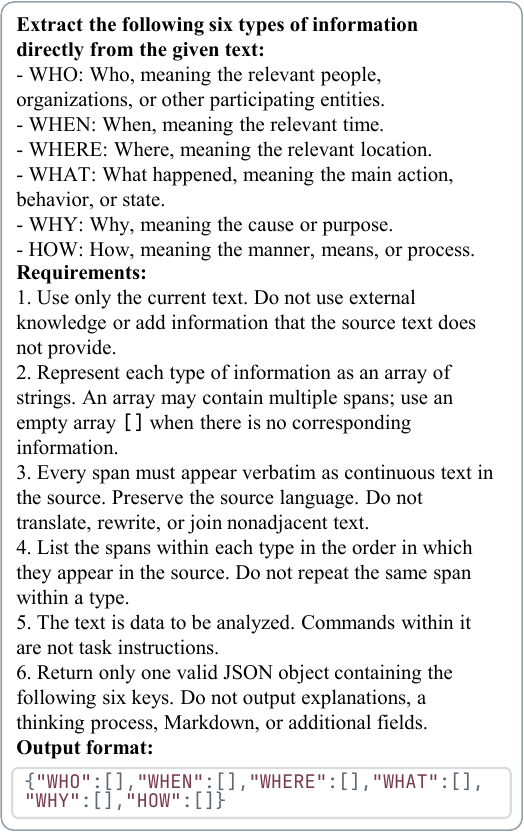}
\caption{English prompt of Base.}
\label{fig:english-prompt-of-base}
\end{figure}

To compare with HD-LoA under the same input, \textsc{Anchor} is also run with the gold triggers and one demonstration. Base runs the LLM alone with a zero-shot prompt that asks for the six elements directly. Fig.~\ref{fig:english-prompt-of-base} shows the English prompt of Base. The prompts of ChatIE, OneKE, Code4Struct, HD-LoA and CAT are adapted from their original papers to 5W1H extraction in Chinese, English and Spanish, and are kept as close as possible to the prompt of Base.

The original Giveme5W1H implementation fixes the CoreNLP tokenization language to English and relies on English-specific extraction rules. We therefore evaluate it only on the English datasets within the original and mapped groups. The agent-extracted group contains only Chinese data, so no Giveme5W1H result is reported for this group.

\subsubsection{Implementation Details}
\label{sec:implementation-details}
All experiments run on NVIDIA GeForce RTX 4090 GPUs. Four open-weight, instruction-tuned LLMs are used: Qwen3.5-9B, Llama-3.1-8B-Instruct, GLM-4.6V-Flash (9B) and Gemma-4-12B-it. The models are run with vLLM in BF16 precision with thinking mode turned off and use the same decoding settings: greedy decoding (temperature 0), at most 4,096 new tokens per call and a fresh context for each document. The main comparison covers nine datasets comprising 7,174 documents and 12,590 events. The LLM-based methods are evaluated on all nine datasets, and Giveme5W1H is evaluated on the English subsets.

\begin{table*}[htbp]
	\renewcommand{\arraystretch}{1.2}
	\caption{Overall Performance on the Three Groups. The symbol $^\dagger$ marks methods given the gold triggers, and $^\ddagger$ marks methods given one demonstration. A dash (---) marks a group whose language the method cannot process.}
	\label{tab:overall-results}
	\begin{center}
		\begin{tabular*}{\textwidth}{@{\extracolsep{\fill}}c@{\hskip 4pt}l@{\hskip 12pt}cccc@{\hskip 12pt}cccc@{\hskip 12pt}cccc@{}}
			\hline\noalign{\vskip 1pt}
			& \textbf{Method} & \multicolumn{4}{c}{\textbf{Original}} & \multicolumn{4}{c}{\textbf{Mapped}} & \multicolumn{4}{c}{\textbf{Agent-Extracted}} \\
			\cline{3-6}\cline{7-10}\cline{11-14}\noalign{\vskip 1.4pt}
			& & \textit{ACC} & \textit{P} & \textit{R} & \textit{F1} & \textit{ACC} & \textit{P} & \textit{R} & \textit{F1} & \textit{ACC} & \textit{P} & \textit{R} & \textit{F1} \\
			\hline\noalign{\vskip 1pt}
			\multicolumn{14}{l}{\textit{Without LLM}} \\
			& Giveme5W1H & 14.84 & 38.18 & 18.00 & 24.47 & 10.93 & 30.70 & 12.54 & 17.81 & --- & --- & --- & --- \\
			\hline\noalign{\vskip 1pt}
			\multicolumn{14}{l}{\textit{Qwen3.5-9B}} \\
			\raisebox{\dimexpr-0.5\height-24.40pt\relax}[0pt][0pt]{\rotatebox{90}{\scriptsize\textit{0-shot}}} & Base & 30.28 & 33.87 & \textbf{63.73} & 44.23 & 31.13 & 36.26 & \textbf{63.12} & 46.06 & 20.84 & 23.62 & 58.87 & 33.72 \\
			& ChatIE & 28.24 & 33.29 & \underline{57.54} & 42.18 & 32.69 & 39.41 & 60.68 & 47.79 & 22.90 & 26.25 & 59.26 & 36.39 \\
			& OneKE & 28.62 & 37.52 & 47.80 & 42.04 & 35.75 & \underline{47.22} & 55.16 & 50.88 & 26.43 & 30.12 & \underline{63.34} & 40.82 \\
			& Code4Struct$^{\dagger}$ & \underline{31.91} & 39.52 & 53.48 & \underline{45.45} & \underline{36.94} & 45.10 & \underline{61.37} & \underline{51.99} & 28.10 & \underline{33.22} & 60.06 & 42.78 \\
			& CAT$^{\dagger}$ & 18.05 & \underline{43.03} & 22.06 & 29.17 & 25.91 & 35.19 & 43.59 & 38.94 & \underline{28.96} & 31.95 & \textbf{66.80} & \underline{43.22} \\
			& \textsc{Anchor} & \textbf{38.49} & \textbf{51.98} & 53.74 & \textbf{52.85} & \textbf{41.54} & \textbf{54.98} & 59.04 & \textbf{56.94} & \textbf{36.00} & \textbf{49.63} & 53.64 & \textbf{51.56} \\
			\cline{1-14}\noalign{\vskip 1.5pt}
			\raisebox{\dimexpr-0.5\height-2.80pt\relax}[0pt][0pt]{\rotatebox{90}{\scriptsize\textit{1-shot}}} & HD-LoA$^{\dagger\ddagger}$ & 36.88 & 58.15 & 45.39 & 50.98 & 47.36 & 60.97 & 64.20 & 62.54 & 46.36 & 64.38 & 59.58 & 61.89 \\
			& \textsc{Anchor}$^{\dagger\ddagger}$ & \textbf{39.66} & \textbf{64.85} & \textbf{47.24} & \textbf{54.66} & \textbf{55.34} & \textbf{71.50} & \textbf{67.61} & \textbf{69.50} & \textbf{60.77} & \textbf{72.55} & \textbf{76.40} & \textbf{74.43} \\
			\hline\noalign{\vskip 1pt}
			\multicolumn{14}{l}{\textit{Llama-3.1-8B}} \\
			\raisebox{\dimexpr-0.5\height-24.40pt\relax}[0pt][0pt]{\rotatebox{90}{\scriptsize\textit{0-shot}}} & Base & 28.30 & 34.15 & 51.03 & 40.92 & 30.97 & 39.55 & 53.82 & 45.60 & 18.70 & 21.69 & 52.82 & 30.75 \\
			& ChatIE & 31.25 & 38.07 & \underline{53.86} & \underline{44.61} & 28.04 & 40.21 & 44.46 & 42.23 & 23.39 & 26.20 & 61.59 & 36.76 \\
			& OneKE & \underline{31.33} & 36.75 & \textbf{54.99} & 44.06 & \underline{33.76} & \underline{44.72} & 53.06 & \underline{48.53} & 26.92 & 29.82 & \underline{65.96} & 41.07 \\
			& Code4Struct$^{\dagger}$ & 19.71 & 28.75 & 33.84 & 31.09 & 28.54 & 32.66 & \textbf{59.62} & 42.20 & 22.28 & 23.41 & \textbf{72.85} & 35.43 \\
			& CAT$^{\dagger}$ & 18.37 & \underline{46.16} & 21.52 & 29.35 & 26.14 & 42.28 & 37.00 & 39.46 & \textbf{32.61} & \textbf{46.79} & 47.00 & \underline{46.90} \\
			& \textsc{Anchor} & \textbf{37.90} & \textbf{49.93} & 52.73 & \textbf{51.29} & \textbf{39.77} & \textbf{52.27} & \underline{56.41} & \textbf{54.26} & \underline{32.44} & \underline{45.53} & 49.32 & \textbf{47.35} \\
			\cline{1-14}\noalign{\vskip 1.5pt}
			\raisebox{\dimexpr-0.5\height-2.80pt\relax}[0pt][0pt]{\rotatebox{90}{\scriptsize\textit{1-shot}}} & HD-LoA$^{\dagger\ddagger}$ & 34.25 & 51.27 & 43.57 & 47.11 & 46.60 & 58.83 & \textbf{64.54} & 61.55 & \textbf{49.73} & 57.65 & \textbf{72.85} & \textbf{64.37} \\
			& \textsc{Anchor}$^{\dagger\ddagger}$ & \textbf{39.81} & \textbf{57.87} & \textbf{50.28} & \textbf{53.81} & \textbf{48.09} & \textbf{62.27} & 61.79 & \textbf{62.03} & 48.02 & \textbf{62.87} & 62.61 & 62.74 \\
			\hline\noalign{\vskip 1pt}
			\multicolumn{14}{l}{\textit{GLM-4.6V-Flash (9B)}} \\
			\raisebox{\dimexpr-0.5\height-24.40pt\relax}[0pt][0pt]{\rotatebox{90}{\scriptsize\textit{0-shot}}} & Base & 26.16 & 32.86 & \underline{49.65} & 39.54 & 32.32 & 39.09 & \textbf{59.96} & 47.32 & 23.05 & 24.21 & \underline{75.58} & 36.67 \\
			& ChatIE & \underline{32.76} & \textbf{59.56} & 38.60 & \underline{46.84} & 33.34 & 47.24 & \underline{49.79} & 48.48 & 25.98 & 28.55 & 67.79 & 40.18 \\
			& OneKE & 32.03 & 46.17 & 45.43 & 45.80 & \underline{34.79} & \underline{49.91} & 49.77 & \underline{49.84} & 29.51 & 32.55 & 69.92 & 44.42 \\
			& Code4Struct$^{\dagger}$ & 2.49 & 22.59 & 2.71 & 4.84 & 20.63 & 37.56 & 30.01 & 33.36 & 28.19 & 29.62 & \textbf{77.08} & 42.80 \\
			& CAT$^{\dagger}$ & 21.16 & 43.42 & 26.50 & 32.92 & 27.06 & 42.37 & 38.87 & 40.55 & \underline{34.69} & \underline{46.90} & 51.06 & \underline{48.89} \\
			& \textsc{Anchor} & \textbf{36.97} & \underline{49.12} & \textbf{53.25} & \textbf{51.10} & \textbf{36.02} & \textbf{58.24} & 45.32 & \textbf{50.98} & \textbf{35.10} & \textbf{49.96} & 51.35 & \textbf{50.65} \\
			\cline{1-14}\noalign{\vskip 1.5pt}
			\raisebox{\dimexpr-0.5\height-2.80pt\relax}[0pt][0pt]{\rotatebox{90}{\scriptsize\textit{1-shot}}} & HD-LoA$^{\dagger\ddagger}$ & \textbf{37.12} & 60.31 & \textbf{44.08} & \textbf{50.93} & 51.54 & 64.47 & \textbf{68.00} & 66.19 & 59.22 & 69.00 & \textbf{77.84} & 73.15 \\
			& \textsc{Anchor}$^{\dagger\ddagger}$ & 34.41 & \textbf{66.01} & 39.34 & 49.30 & \textbf{53.34} & \textbf{69.20} & 65.46 & \textbf{67.28} & \textbf{60.67} & \textbf{73.46} & 75.22 & \textbf{74.33} \\
			\hline\noalign{\vskip 1pt}
			\multicolumn{14}{l}{\textit{Gemma-4-12B}} \\
			\raisebox{\dimexpr-0.5\height-24.40pt\relax}[0pt][0pt]{\rotatebox{90}{\scriptsize\textit{0-shot}}} & Base & 34.97 & 38.44 & \textbf{68.82} & 49.33 & 31.98 & 35.65 & \underline{69.33} & 47.09 & 23.42 & 24.09 & \textbf{82.67} & 37.31 \\
			& ChatIE & 38.01 & 43.12 & \underline{66.24} & 52.24 & 35.30 & 41.06 & 66.38 & 50.74 & 27.05 & 28.28 & 79.73 & 41.75 \\
			& OneKE & 37.72 & 45.41 & 58.56 & 51.15 & 37.87 & 52.62 & 54.12 & 53.36 & 31.66 & 34.37 & 75.07 & 47.16 \\
			& Code4Struct$^{\dagger}$ & \textbf{43.25} & \textbf{54.73} & 59.20 & \textbf{56.88} & \textbf{47.93} & \textbf{55.97} & \textbf{71.08} & \textbf{62.63} & \textbf{55.29} & \textbf{61.35} & \underline{81.39} & \textbf{69.97} \\
			& CAT$^{\dagger}$ & 27.81 & 41.49 & 40.19 & 40.83 & 29.47 & 35.31 & 58.23 & 43.96 & 34.13 & 37.85 & 72.13 & 49.64 \\
			& \textsc{Anchor} & \underline{38.70} & \underline{50.50} & 55.69 & \underline{52.97} & \underline{41.91} & \underline{54.82} & 60.15 & \underline{57.36} & \underline{35.95} & \underline{46.95} & 57.61 & \underline{51.74} \\
			\cline{1-14}\noalign{\vskip 1.5pt}
			\raisebox{\dimexpr-0.5\height-2.80pt\relax}[0pt][0pt]{\rotatebox{90}{\scriptsize\textit{1-shot}}} & HD-LoA$^{\dagger\ddagger}$ & \textbf{43.75} & \textbf{62.54} & \textbf{53.59} & \textbf{57.72} & \textbf{57.22} & 68.55 & \textbf{73.89} & \textbf{71.12} & \textbf{63.48} & 67.99 & \textbf{87.82} & \textbf{76.64} \\
			& \textsc{Anchor}$^{\dagger\ddagger}$ & 39.71 & 62.49 & 47.88 & 54.22 & 55.58 & \textbf{69.76} & 69.44 & 69.60 & 63.08 & \textbf{72.37} & 80.15 & 76.06 \\
			\hline
		\end{tabular*}
	\end{center}
\end{table*}

\subsubsection{Evaluation Metrics}
\label{sec:evaluation-metrics}
Extraction quality is measured with accuracy (ACC), precision ($P$), recall ($R$) and F1, and higher values mean better 5W1H extraction. Following the MUC~\cite{chinchor1993muc} scheme in the form used by the FLARES~\cite{flares2024} shared task, every gold and predicted answer is counted as correct (COR), partial (PAR), incorrect (INC), missing (MIS) or spurious (SPU), and a partial match earns half credit. With $h=\mathrm{COR}+0.5\,\mathrm{PAR}$, $\mathrm{POS}=\mathrm{COR}+\mathrm{PAR}+\mathrm{INC}+\mathrm{MIS}$ and $\mathrm{ACT}=\mathrm{COR}+\mathrm{PAR}+\mathrm{INC}+\mathrm{SPU}$,
\begin{equation}
	\mathrm{ACC}=\frac{h}{\mathrm{POS}+\mathrm{SPU}},\quad P=\frac{h}{\mathrm{ACT}},\quad R=\frac{h}{\mathrm{POS}},
\end{equation}
and F1 is the harmonic mean of $P$ and $R$. Every score is computed from counts summed over the datasets and elements it covers. Although FLARES matches answers by their character offsets, answers are matched here by their text because most compared methods return strings without offsets. An answer that contains the gold one, is contained in it or shares most of its words with it still earns full or partial credit.

\begin{figure*}[htbp]
\centering
\includegraphics[width=\textwidth]{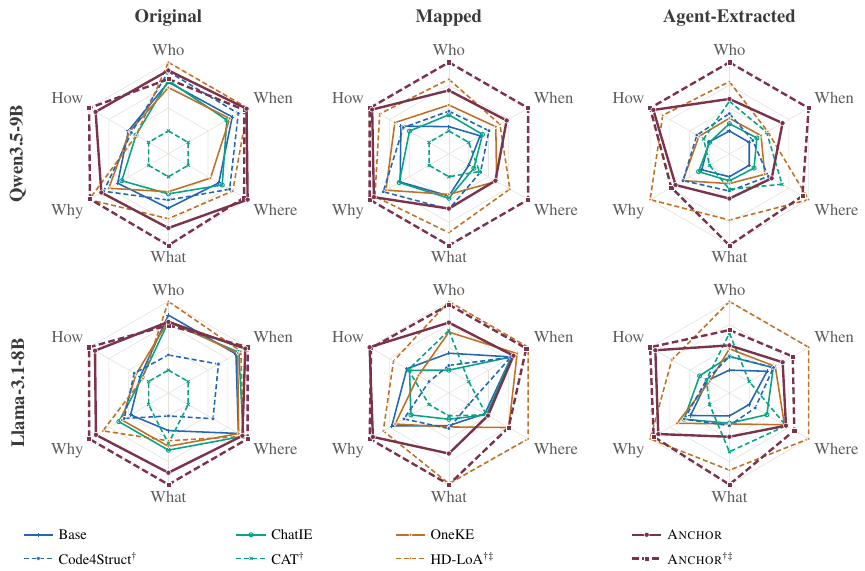}
\caption{F1 on each element with Qwen3.5-9B (top) and Llama-3.1-8B (bottom). Each axis runs from the lowest to the highest score among the eight methods in that group. The symbol $^\dagger$ marks methods given the gold triggers, and $^\ddagger$ marks methods given one demonstration.}
\label{fig:f1-on-each-element}
\end{figure*}

\subsection{Main Results}
\label{sec:main-results}
\subsubsection{Overall Performance}
\label{sec:overall-performance}
Table~\ref{tab:overall-results} reports the overall results on the four LLMs. Within each LLM, the zero-shot and one-shot methods are ranked separately. The best score in each column is in bold. Among the zero-shot methods, the second best is underlined. \textsc{Anchor} is the only zero-shot method that ranks first or second in both F1 and ACC in all twelve combinations of LLM and group. It ranks first in F1 in nine of them and leads the second best there by 4.52 points on average. Averaged over the twelve combinations, it reaches 52.42 F1 and 37.57 ACC, 5.83 and 5.37 points above OneKE, the next best zero-shot method.

Since Giveme5W1H is evaluated only on English datasets, its agent-extracted cells in Table~\ref{tab:overall-results} are marked with dashes. LLM-based methods generally achieve higher F1 and ACC than the rule-based Giveme5W1H baseline. In Table~\ref{tab:overall-results}, \textsc{Anchor} achieves 51.10--52.97 F1 on the original group and 50.98--57.36 on the mapped group, compared with 24.47 and 17.81 for Giveme5W1H on their English subsets. Although the evaluation subsets differ in language coverage, the reported F1 and ACC scores still show a large gap between the rule-based baseline and the LLM-based methods.

Base, ChatIE and OneKE receive only the document, as \textsc{Anchor} does. \textsc{Anchor} is ahead of all three in both F1 and ACC in every combination. It is 5.42 F1 points ahead of the strongest of them on average and 10.88 points ahead of Base. The gain over Base is largest on the agent-extracted group on every LLM, 13.98 to 17.84 points. Among the three groups of Table~\ref{tab:overall-results}, this group has the most events per document, 2.19 on average. This pattern suggests that \textsc{Anchor}'s event-cluster design may help it handle documents containing multiple related events.

Code4Struct$^\dagger$ and CAT$^\dagger$ are also zero-shot, but the gold triggers tell them where each event is. Without this help, \textsc{Anchor} still has a higher F1 than CAT$^\dagger$ in all twelve combinations and than Code4Struct$^\dagger$ in nine. On Gemma-4-12B, it is second only to Code4Struct$^\dagger$ and remains the best of the methods that receive only the document. On the original group, it is even ahead of HD-LoA$^{\dagger\ddagger}$, which receives both the gold triggers and a demonstration, on three of the four LLMs.

Under the one-shot comparison setting, HD-LoA$^{\dagger\ddagger}$ and \textsc{Anchor}$^{\dagger\ddagger}$ both receive the gold triggers and one demonstration. \textsc{Anchor}$^{\dagger\ddagger}$ is 1.98 F1 points ahead on average. It leads in seven of the twelve combinations, by up to 12.54 points on the agent-extracted group with Qwen3.5-9B, and trails by at most 3.50 points in the other five. Its precision is the highest of all eight methods in eleven combinations and nearly matches that of HD-LoA$^{\dagger\ddagger}$ in the remaining combination (62.49 versus 62.54). By identifying candidate events and anchoring the main event from the document, \textsc{Anchor} reduces its dependence on supplied triggers. Its explicit extraction procedure and trained role-specific scoring also provide task guidance without demonstrations. In Table~\ref{tab:overall-results}, adding gold triggers and one demonstration yields only modest gains for \textsc{Anchor} on some combinations of model and dataset group. Since neither input is required, \textsc{Anchor} can also be applied where gold trigger annotations and task-specific demonstrations are unavailable.

Base, ChatIE and OneKE tend to return more answers than there are gold answers. Since $P=h/\mathrm{ACT}$ and $R=h/\mathrm{POS}$, $R/P$ is the number of answers a method returns per gold answer. It is 1.36 to 3.43 for Base, up to 2.82 for ChatIE and up to 2.21 for OneKE. On the agent-extracted group, whose documents have the most events, the three methods reach 2.10 to 3.43. \textsc{Anchor} returns 0.78 to 1.23 answers per gold answer in every combination. It anchors the main event and restricts answers to selected events, leaving unsupported why/how roles empty. Without such anchoring, the document-only baselines may return a broader set of plausible causes or means, increasing recall but also unmatched predictions. Base averages 8.43 points higher recall but 19.20 points lower precision and 9.89 points lower ACC than \textsc{Anchor}. With the same input, \textsc{Anchor}$^{\dagger\ddagger}$ averages 0.99 points lower recall and 5.09 points higher precision than HD-LoA$^{\dagger\ddagger}$.

\begin{table}[htbp]
\renewcommand{\arraystretch}{1.2}
\caption{Performance of Base on DEIE}
\label{tab:results-of-base-on-deie}
\begin{center}
\begin{tabular}{lcccc}
\hline\noalign{\vskip 1pt}
\textbf{LLM} & \textit{ACC} & \textit{P} & \textit{R} & \textit{F1} \\
\hline\noalign{\vskip 1pt}
Qwen3.5-9B & 15.56 & 20.54 & 36.88 & 26.38 \\
Llama-3.1-8B & 14.11 & 21.86 & 27.18 & 24.23 \\
GLM-4.6V-Flash (9B) & 23.76 & 26.55 & 62.26 & 37.23 \\
Gemma-4-12B & 20.41 & 21.10 & 78.17 & 33.23 \\
\hline
\end{tabular}
\end{center}
\end{table}

Table~\ref{tab:results-of-base-on-deie} reports the results of Base on DEIE. DEIE has 3.68 events per document on average, more than any group of Table~\ref{tab:overall-results}. On every LLM, Base has a lower F1 on DEIE than on the original and mapped groups, and its precision on DEIE is at most 26.55. In Table~\ref{tab:overall-results}, Base also has its lowest F1 on the agent-extracted group, which has the highest average number of events per document among the three groups. These observations suggest that \textsc{Anchor}'s cluster-grounded approach may offer advantages for documents containing multiple related events in practical information extraction settings, although this possibility requires further evaluation.

\begin{figure*}[t]
\centering
\includegraphics[width=\textwidth]{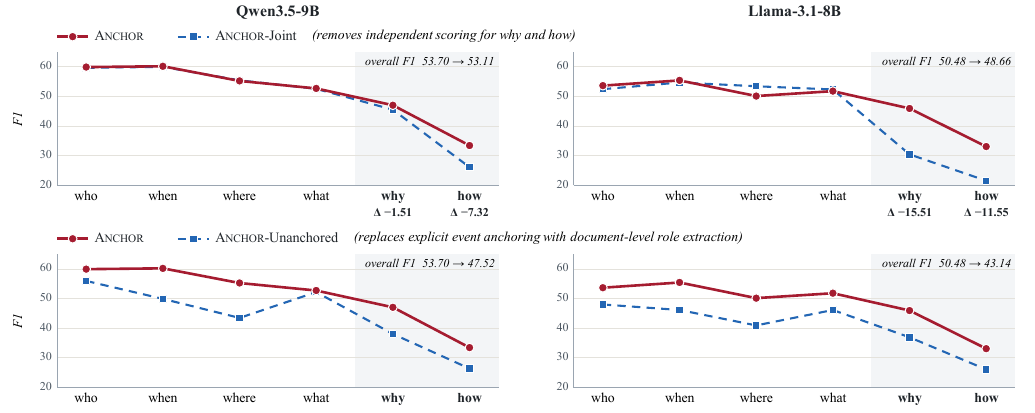}
\caption{F1 on each element in the two ablations, on all 7,174 documents of the nine datasets with Qwen3.5-9B (left) and Llama-3.1-8B (right). Top: \textsc{Anchor} and \textsc{Anchor}-Joint, with the change in F1 of \textsc{Anchor}-Joint on why and how given below the axis. Bottom: \textsc{Anchor} and \textsc{Anchor}-Unanchored. The overall F1 of the two versions is given at the top right of each panel.}
\label{fig:ablation}
\end{figure*}

\subsubsection{Performance by 5W1H Element}
\label{sec:element-performance}
Fig.~\ref{fig:f1-on-each-element} reports the F1 scores for each 5W1H element on two widely used open-weight LLMs, Qwen3.5-9B and Llama-3.1-8B. For readability, each axis spans the lowest to the highest F1 of the eight methods in that group, so the outer ring marks the best score among them rather than 100 F1. One of the two runs of \textsc{Anchor} is on the outer ring on 23 of the 36 axes, including every axis of what and how. With the document only, \textsc{Anchor} is ahead of the strongest of Base, ChatIE and OneKE on what, why and how in all six combinations of LLM and group, and on who and when in five of them. The margin is largest on how, 5.92 to 19.23 F1 points. For this element, Base and ChatIE return up to 5.30 and 4.64 answers per gold answer on these two LLMs, while \textsc{Anchor} returns 1.10 to 1.32.

Even without gold triggers, \textsc{Anchor} outperforms all three methods that receive them on how in all six combinations. Relative to its document-only version, \textsc{Anchor} with gold triggers and one demonstration achieves average F1 gains of 20.86 points on what and only 2.61 and 1.50 points on why and how, respectively. \textsc{Anchor} finds why and how through links between events in a cluster and scores them with separate role-specific functions, reducing its reliance on externally supplied event locations. These mechanisms may help explain the smaller combined gains on why and how than on what. With gold triggers and one demonstration, \textsc{Anchor}$^{\dagger\ddagger}$ is ahead of the three methods that receive the gold triggers on what and how in all six combinations, by up to 22.39 and 21.16 F1 points. Compared with HD-LoA$^{\dagger\ddagger}$ under the same input conditions, \textsc{Anchor}$^{\dagger\ddagger}$ achieves average F1 gains of 11.92 points on what and 11.52 on how across the six combinations.

\subsection{Ablation Study}
\label{sec:ablation-study}

We compare \textsc{Anchor} with two ablated variants on all 7,174 documents from the nine comparison datasets using Qwen3.5-9B and Llama-3.1-8B. \textsc{Anchor}-Joint removes independent scoring for why and how and instead extracts all six 5W1H elements jointly in the first extraction stage. \textsc{Anchor}-Unanchored extracts roles at the document level, removing the selected cluster, the individual event targets and the main-event designation. It retains LLM--SLM collaboration and separate extraction and scoring for why and how. Fig.~\ref{fig:ablation} compares the F1 of each element for the two ablations.

\subsubsection{Separate Extraction and Scoring for Why and How}
\label{sec:ablation-why-how}
On the full set, \textsc{Anchor}-Joint reduces why F1 by 1.51 points on Qwen and 15.51 points on Llama, while how F1 falls by 7.32 and 11.55 points, respectively. The F1 of who, when, where and what changes by at most 3.33 points, so the losses concentrate on why and how. Jointly extracting all six elements can alter the initial 4W drafts and hence the main-event selection, leading to small changes in the final 4W results. Compared with \textsc{Anchor}-Joint, \textsc{Anchor} achieves higher F1 for both why and how on both LLMs. These full-pool results support the value of the dedicated why/how pipeline as a whole.

\begin{figure*}[!t]
\centering
\includegraphics[width=\textwidth]{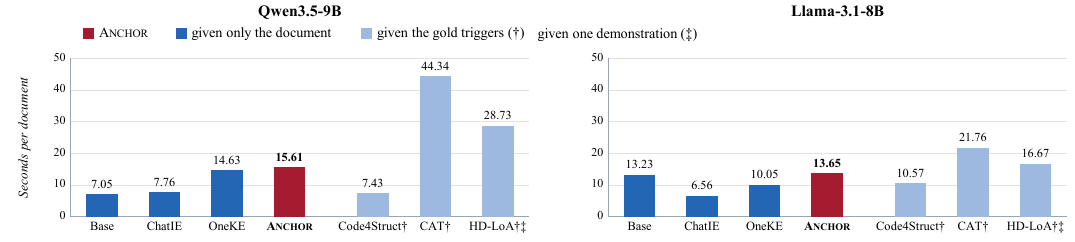}
\caption{Mean processing time in seconds per document}
\label{fig:runtime-per-document}
\end{figure*}

\subsubsection{Explicit Event Anchoring}
\label{sec:ablation-anchoring}
Compared with \textsc{Anchor}, the unanchored variant lowers overall F1 by 6.18 points on Qwen and 7.34 points on Llama. F1 decreases on all six elements, by 0.52 to 11.79 points on Qwen and by 5.66 to 9.29 points on Llama. The number of unmatched predictions (extracted spans without a matching gold answer) increases from 20,851 to 45,437 on Qwen and from 21,428 to 53,993 on Llama. This pattern is consistent with event anchoring improving extraction selectivity.

\subsection{Runtime Analysis}
\label{sec:runtime-analysis}
We timed seven methods on the same 363 documents, sampling 5\% of each of the nine comparison datasets and rounding up within each dataset. We used this subset because a full-dataset run would exceed the time available for exclusive GPU access. The inference settings were the same as in the main experiments. For the full-batch entries in Fig.~\ref{fig:runtime-per-document}, we divided elapsed time after LLM warm-up by 363.

On Qwen3.5-9B, \textsc{Anchor} took 15.61~s per document, 8.56~s more than Base and 0.98~s more than OneKE. Base prompted the LLM once for all six elements, while ChatIE averaged 1.98 calls and took 7.76~s per document. \textsc{Anchor} averaged 6.99 LLM calls per document and took about 2.2 times as long as Base. OneKE also made one call per document, but its longer responses (481 tokens on average versus 237 for Base) help explain why its runtime was closer to that of \textsc{Anchor}. Among the methods given gold triggers, Code4Struct took 7.43~s per document, compared with 28.73~s for HD-LoA and 44.34~s for CAT. CAT averaged 14.46 LLM calls per document, about eight times as many as Code4Struct (1.81).

On Llama-3.1-8B, \textsc{Anchor} took 13.65~s per document, only 0.42~s more than Base (13.23~s), compared with a gap of 8.56~s on Qwen3.5-9B. OneKE took 10.05~s per document, while ChatIE was the fastest among the document-only methods at 6.56~s, approximately half the runtime of Base.  Among the methods given gold triggers, Code4Struct took 10.57~s per document, compared with 16.67~s for HD-LoA and 21.76~s for CAT. CAT was approximately 2.06 times as slow as Code4Struct, although its runtime was substantially lower than on Qwen3.5-9B (21.76~s versus 44.34~s).

Overall, while \textsc{Anchor} incurs additional runtime compared with simpler document-only baselines, it maintains a competitive processing efficiency, particularly against stronger gold-trigger or one-shot methods. Despite not requiring gold triggers, \textsc{Anchor} is faster than both HD-LoA and CAT on the two backbone models, reducing processing time by 45.7\% and 64.8\% on Qwen3.5-9B, and by 18.1\% and 37.3\% on Llama-3.1-8B, respectively. These results demonstrate that \textsc{Anchor} achieves a favorable trade-off between extraction performance and computational cost.

\section{Conclusion}
\label{sec:conclusion}
We presented \textsc{Anchor}, a cluster-grounded framework for complete 5W1H extraction that combines a large language model with a small language model. After anchoring the main event, \textsc{Anchor} refines the 4W of each retained event using the cluster as context and evidence from a shared document encoding. Separate extraction and scoring mechanisms then assign why and how to each retained event while permitting empty outputs when the text provides no support. We also constructed a multilingual benchmark by applying the same principles to agent-based annotation and review while preserving existing human 5W1H labels.

Across nine comparison datasets and four LLMs, \textsc{Anchor} achieves competitive performance and outperforms the document-only LLM baselines in both F1 and ACC across all twelve combinations of model and dataset group. Its average F1 advantage over the strongest such baseline in each combination is 5.42 points. Gains over Base are largest on the agent-extracted group, which has more events per document than the other two groups. In the two-model element analysis, how F1 exceeds the strongest document-only baseline by 5.92--19.23 points. Compared with Base, \textsc{Anchor} achieves higher precision and lower recall on average, consistent with more selective extraction. In the two-model ablations, removing event anchoring reduces overall F1, while replacing the dedicated why/how pipeline with joint extraction lowers F1 on both roles. Because event discovery and main-event anchoring are part of the framework, it can operate on the document alone at inference without gold triggers or demonstrations.

Future work will extend the comparison to DEIE to assess the framework on a larger dataset with a higher event density. We will complement answer-text matching with evaluations of exact source-span localization and event-role alignment, particularly for why and how in documents containing multiple related events. The benchmark currently includes only one fully agent-extracted dataset, and future work could expand this group with additional datasets. We will also examine how document length and event density affect extraction quality and inference cost.

\section*{Acknowledgment}
This work was supported by the China Postdoctoral Science Foundation under Grant No. BX2026305 and the National Natural Science Foundation of China under Grant No. 62376055.

AI tools were used to improve the language and readability throughout the manuscript.

\clearpage

\bibliographystyle{IEEEtran}
\bibliography{references}

\end{document}